\documentclass[a4paper]{article}

\usepackage{geometry}
\usepackage{gensymb}
\usepackage{graphicx, hyperref, setspace, amsmath, amssymb, titlesec, fancyhdr, multicol, parskip, indentfirst, etoolbox, caption, cite, xcolor, float}

\titleformat{\section}{\centering\large\scshape}{\thesection}{1em}{}
\titleformat{\subsection}{\normalsize\bfseries}{\thesubsection.}{1em}{}
\titlespacing{\section}{0pt}{6pt}{6pt}
\titlespacing{\subsection}{0pt}{6pt}{6pt}
\titlespacing{\subsubsection}{0pt}{6pt}{6pt}
\title{
    \textbf{MultiRetNet: A Multimodal Vision Model and Deferral System for Staging Diabetic Retinopathy}
}

\date{} 
\renewcommand{\thesection}{\Roman{section}.}
\renewcommand{\thesubsection}{\textit{\Alph{subsection}.}}
\renewcommand{\thesubsubsection}{\textit{\arabic{subsubsection}.}}

\titleformat{\subsection}{\normalfont\large\itshape}{\thesubsection}{1em}{}
\titleformat{\subsubsection}{\normalfont\itshape}{\thesubsubsection}{1em}{}

\begin{document}

\maketitle
\vspace{-1.5cm}


\begin{multicols}{2}
    \centering

    \textbf{Jeannie She\textsuperscript{*\;1}}\\
    \textit{jeanshe@mit.edu}\\
    \vfill

    \columnbreak

    \textbf{Katie Spivakovsky\textsuperscript{*\;1\;2}}\\
    \textit{kspiv@mit.edu}\\
    \vfill
\end{multicols}

\footnotetext{
\noindent 
\vspace*{\fill}
\\
\noindent\rule{\linewidth}{0.4pt} \\[4pt] 
\textsuperscript{*}Equal contribution\\
\textsuperscript{1}Department of Electrical Engineering and Computer Science, MIT\\
\textsuperscript{2}Department of Biological Engineering, MIT}

\singlespacing
\setlength{\parskip}{6pt}
\setlength{\parindent}{0.5cm}

\begin{multicols}{2}
\setlength{\columnsep}{0.5cm}

\section{Abstract}
Diabetic retinopathy (DR) is a leading cause of preventable blindness, affecting over 100 million people worldwide. In the United States, individuals from lower-income communities face a higher risk of progressing to advanced stages before diagnosis, largely due to limited access to screening \cite{lu_divergent_2016}. Comorbid conditions further accelerate disease progression.
We propose MultiRetNet, a novel pipeline combining retinal imaging, socioeconomic factors, and comorbidity profiles to improve DR staging accuracy, integrated with a clinical deferral system for a clinical human-in-the-loop implementation. We experiment with three multimodal fusion methods and identify fusion through a fully connected layer as the most versatile methodology. We synthesize adversarial, low-quality images and use contrastive learning to train the deferral system, guiding the model to identify out-of-distribution samples that warrant clinician review. By maintaining diagnostic accuracy on suboptimal images and integrating critical health data, our system can improve early detection, particularly in underserved populations where advanced DR is often first identified. This approach may reduce healthcare costs, increase early detection rates, and address disparities in access to care, promoting healthcare equity.

\section{Introduction}

Diabetic retinopathy (DR) represents one of the leading causes of preventable blindness worldwide; in the United States alone, nearly 10 million individuals suffer from this sight-threatening complication of diabetes, with disproportionately high prevalence among socioeconomically disadvantaged populations \cite{cdc_vehss_2025}. The presence of comorbidities significantly impacts disease progression, with hypertension increasing the risk of severe DR by 2-3 fold and chronic kidney disease accelerating progression by up to 4 fold \cite{uk_prospective_diabetes_study_group_tight_1998}, \cite{wong_increased_2016}. As of 2025, the first AI-driven, FDA-approved tool for diagnosing DR (IDx-DR) is in use globally \cite{niederhauser_digital_2020}. However, it is restricted to high-quality images captured with specialized digital fundus cameras and cannot process poor-quality images, such as blurry ones. A major challenge for socioeconomically disadvantaged populations is the poor quality of fundus images, which can delay DR diagnosis. In fact, 14\% of imaging cases require patients to return for another session, prolonging the time until treatment is initiated \cite{tufail_automated_2017}.

\begin{figure}[H]
    \centering    
    \includegraphics[trim={0 0 0 10},clip,width=\linewidth]{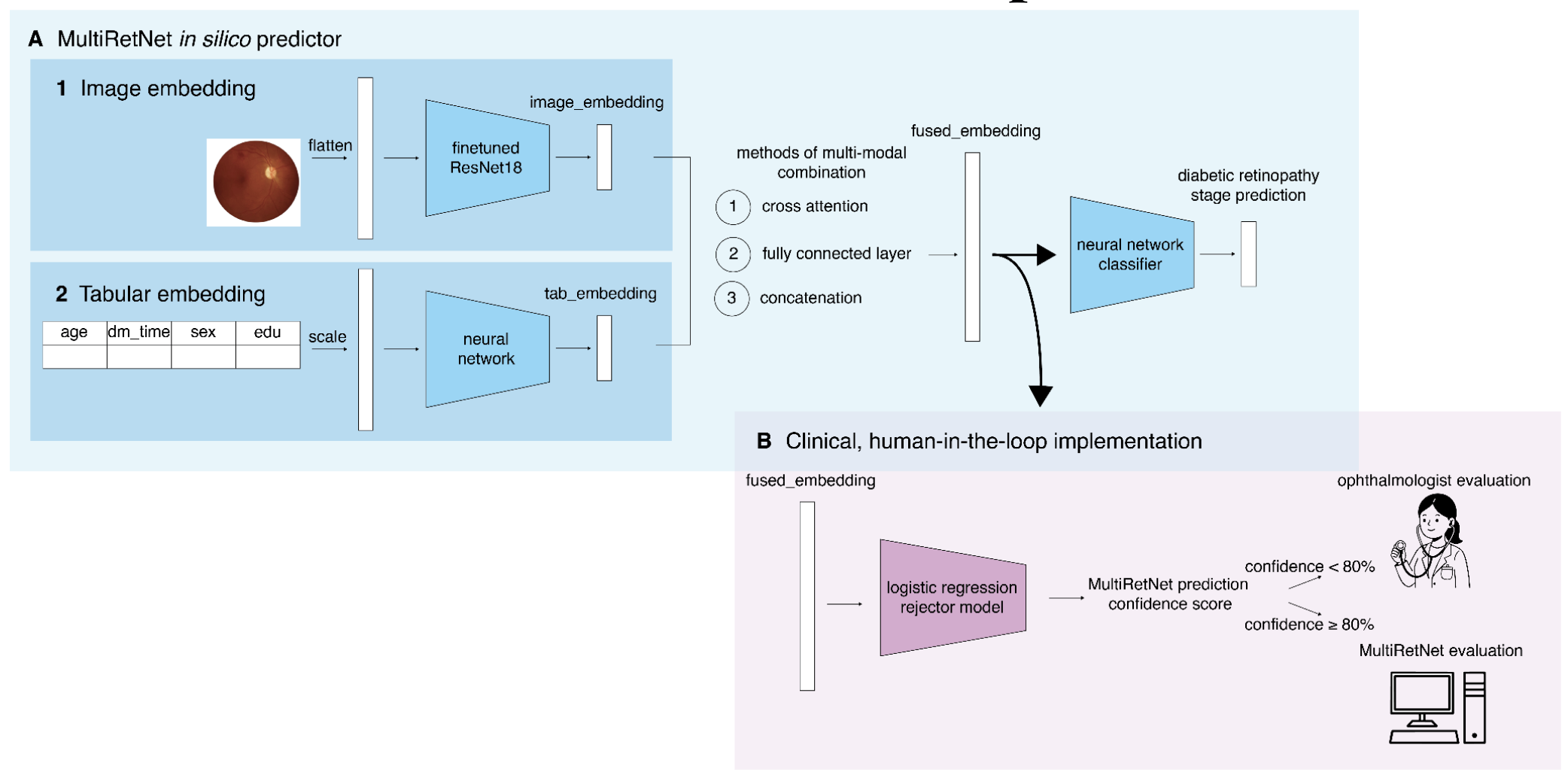}
    \caption{MultiRetNet pipeline with an in silico multimodal predictor that feeds into a deferral system incorporating human decision making.}
    \label{fig:pipeline}
\end{figure}

We propose a pipeline schematized in Figure \ref{fig:pipeline} that incorporates retinal imaging data with socioeconomic determinants of health and comorbidity profiles to more accurately stage DR. We train a deferral system on adversarial images that are blurry or unusually rotated to predict when out-of-distribution samples should be reviewed by clinicians instead of MultiRetNet. Our computer vision pipeline represents a significant advancement in DR screening; implementation of this system could reduce the estimated \$93,000 lifetime cost per case of blindness due to DR \cite{javitt_preventive_1994} while potentially increasing early detection rates in underserved communities, thus also bringing awareness to the critical need for healthcare equity irrespective of socioeconomic factors.

\end{multicols}

\begin{multicols}{2}

\section{Related Work}
Extensive deep learning models predicting DR have been produced, such as Dai et al. (2021) \cite{dai_deep_2021} who identify the lesions on the retinal image itself and label their medical classification. However, existing models only utilize images as input, disregarding potential related implications of patient demographics and clinical history.
Since the dataset we are using contains both images and tabular features, we are heavily inspired by Li et al. (2024) \cite{li_review_2024}, who discuss different methods of fusing multimodal data and encourage cross-attention in particular to be further explored. We choose to experiment with cross-attention, concatenation, and fully-connected layer-based fusion to determine the most robust architecture for MultiRetNet.
Finally, we are motivated by the mission of human-centered AI; pain points in the DR screening process outlined by Beede et al. (2020) \cite{beede_human-centered_2020} emphasize the need to deeply consider how these models will be deployed at clinics with limited access to internet, time per patient, and clinicians. To that end, we look to the contrastive learning methods described by Chen et al. (2020) \cite{chen_simple_2020} to train our deferral system.

\section{Methods}
\subsection{Data preprocessing}
mBRSET \cite{wu_portable_2025} \cite{nakayama_mbrset_nodate} \cite{goldberger_physiobank_2000} samples labeled with unknown (NaN) DR staging were removed. The remaining NaNs in the \texttt{dm\_time} and \texttt{educational\_level} features (1\% of the patients) were imputed with the mode taken across all samples for that given feature. We kept 4884 images corresponding to 1287 patients as our final dataset. Features were standardized to zero mean and unit variance using the StandardScaler from the \texttt{scikit-learn} library. Images were cropped to 224x224 pixels and normalized using the channel-wise statistics of the ImageNet dataset (mean = [0.485, 0.456, 0.406], std = [0.229, 0.224, 0.225]), consistent with the convention for the pretrained ResNet18. Train and test datasets were generated through a random 80\%/20\% split. 

\subsection{Unimodal baseline models}
\paragraph{ResNet18}
We began with the ResNet18 backbone pretrained on ImageNet as our image classifier. To establish a baseline, we evaluated the out-of-the-box performance of ResNet18 directly on our dataset to understand how well generic ImageNet features transfer to our DR stage classification task. We replaced the final classification layer to predict only five classes to correspond with our five labels. Next, we addressed the extreme class imbalance in our data, where 76.7\% of our data were labeled Class 0. For fine-tuning ResNet18, we utilized frequency weighting in our loss function to ensure the model predicted all labels with equal importance. The resulting class weights used in the cross entropy loss function were: [0.2605, 3.5912, 1.7197, 11.9122, 4.6075], corresponding to the five classification categories.  The model was trained on 80\% of our dataset for 20 epochs using the Adam optimizer (learning rate = 0.001).

\paragraph{Tabular Neural Network}
We designed a simple feed-forward neural network consisting of:
\begin{enumerate}
    \item A fully-connected input layer accepting 17 features and producing 32 outputs with ReLU activation
    \item A fully-connected output layer accepting 32 features and producing 5 logits for the 5 DR stages
\end{enumerate}
The model was trained for 20 epochs using the Adam optimizer (learning rate = 0.001) with loss calculated through cross entropy weighted by each class' abundance in the dataset. Shapley additive explanation (SHAP) values were calculated using the \texttt{SHAP} library to determine the relative impact of each of mBRSET's 17 tabular features on the neural network's output. By balancing two aspects of features—predictive power as quantified by SHAP values and our intuitive sense of clinical relevance—we isolated four critical tabular features for use in future modeling.

\subsection{Multimodal fusion experimentation}
We initialized a ResNet18 model with weights obtained from prior fine-tuning and removed its final two layers to extract 512-dimensional image embeddings. For the tabular data, we selected four features (educational level, sex, time since diabetes diagnosis, and age) based on SHAP analysis and encoded each scalar value using a shared projection layer consisting of a linear transformation followed by layer normalization. This resulted in a set of tabular embeddings with the same dimensionality as the image embeddings. The image and tabular representations were then fused using one of the following three fusion strategies. We wished to explore the performance of methods ranging from most naive (concatenation) to most dynamic and non-linear (cross-attention). MultiRetNet contains a final neural network tasked to learn DR stage classifications from multimodal, fused embeddings. To address class imbalance, we downsampled the dataset to include 82 images per class. We opted for downsampling over inverse frequency weighting in this task for simplicity. To ensure the model still learned generalizable features and avoided overfitting despite the reduced dataset size, we used 5-fold cross-validation with 10 training epochs, cross-entropy loss, and the Adam optimizer (learning rate = 0.001). All ResNet18 weights derived from prior fine-tuning were kept frozen during training, allowing only the weights in the fusion and classifier layers to be updated. We evaluated on a held-out, class balanced test set.

\paragraph{Concatenation}
We applied mean pooling separately to the image and tabular embeddings and concatenated the resulting vectors to form a joint representation.
\paragraph{Fully-connected layer}
Following mean pooling and concatenation, we passed the fused vector through a single fully-connected layer with input and output dimensionality of 1024.
\paragraph{Cross-attention}
We applied bidirectional cross-attention to enable mutual information exchange between modalities. In one direction, the image embeddings serve as queries while the tabular embeddings served as the keys and values; in the other direction, the setup is reversed. The outputs from both attention directions were mean pooled and concatenated to form the final fused representation.

\subsection{Deferral system}
\paragraph{Adversarial image generation}
We randomly selected twenty images from each of the five DR stages in mBRSET. For each image, we applied a series of transformations: random rotation within the range of -30$\degree$ to 30$\degree$, Gaussian blur with a kernel size of 5 pixels, and random color jitter with maximum adjustments of brightness (0.2), contrast (0.2), and saturation (0.2). These transforms are inspired by common difficulties in capturing high-quality fundus images in clinic, namely poor lighting and rushed imaging \cite{beede_human-centered_2020}. This procedure yielded a total of 100 adversarial, class-balanced images.

\paragraph{Contrastive learning}
We developed a contrastive learning framework to distinguish between relatively high-quality mBRSET images and our low-quality adversarial images based on their embedded feature representations, while retaining tabular information. More precisely, the input to the contrastive learning framework is an embedding generated through our multimodal model framework, which contains information both about images and about tabular features. The framework consists of a neural network architecture, a specialized loss function, and visualization techniques to evaluate embedding quality.

We designed a neural network, \texttt{ContrastiveNet}, that maps  embeddings to a new projection space where similar-quality images cluster together while dissimilar-quality images are pushed apart. The network consists of:
\begin{enumerate}
    \item A projection network with: \begin{enumerate}
        \item An input layer accepting 1024-dimensional embeddings
        \item A fully-connected hidden layer with 128 neurons and ReLU activation
        \item A projection layer producing 64-dimensional embeddings
        \item Batch normalization to stabilize training
   \end{enumerate}

    \item A binary classification head with:
    \begin{enumerate}
        \item A linear layer from the projection space to a single output
        \item Sigmoid activation for probability output
   \end{enumerate}
\end{enumerate}

We implemented a custom contrastive loss function that operates on pairwise distances between embeddings in the projection space. The loss contains two components:
\begin{enumerate}
    \item \textbf{Positive pair loss}: Encourages embeddings of the same class (both good or both poor quality) to be close to each other in the projection space.
    \item \textbf{Negative pair loss}: Pushes embeddings of different classes (good vs. poor quality) apart by at least a margin distance $\epsilon$.
\end{enumerate}

\noindent The combined loss is computed as:

$$L = \frac{\sum_{i,j} P_{ij} \cdot d(x_i, x_j)}{\sum_{i,j} P_{ij}} + \frac{\sum_{i,j} N_{ij} \cdot \max(0, \epsilon - d(x_i, x_j))}{\sum_{i,j} N_{ij}}$$

\noindent where $P_{ij}$ is 1 if samples $i$ and $j$ belong to the same class (0 otherwise); $N_{ij}$ is 1 if samples $i$ and $j$ belong to different classes; and $d(x_i, x_j)$ is the Euclidean distance between the projections of samples $i$ and $j$. This loss is combined with binary cross-entropy loss from the classification head to form the final training objective.

The model was trained using the Adam optimizer with initial learning rate = 0.001 and a scheduler that reduced the learning rate by half after 5 epochs without improvement in validation loss. Training proceeded for 100 epochs with a batch size of 64, and 20\% of the data was reserved for validation.

For deployment, we implemented a threshold-based classification system where images with a quality score above 0.8 were classified as good quality, while those below required further assessment and would be deferred to a clinician. To assess the effectiveness of the learned embedding space, we employed t-Distributed Stochastic Neighbor Embedding (t-SNE) to reduce both the original embeddings and the model's projected embeddings to a 2D space while preserving local neighborhood structure.

\section{Results}

\subsection{Unimodal models}
\begin{figure}[H]
    \centering    
    \includegraphics[width=0.825\linewidth]{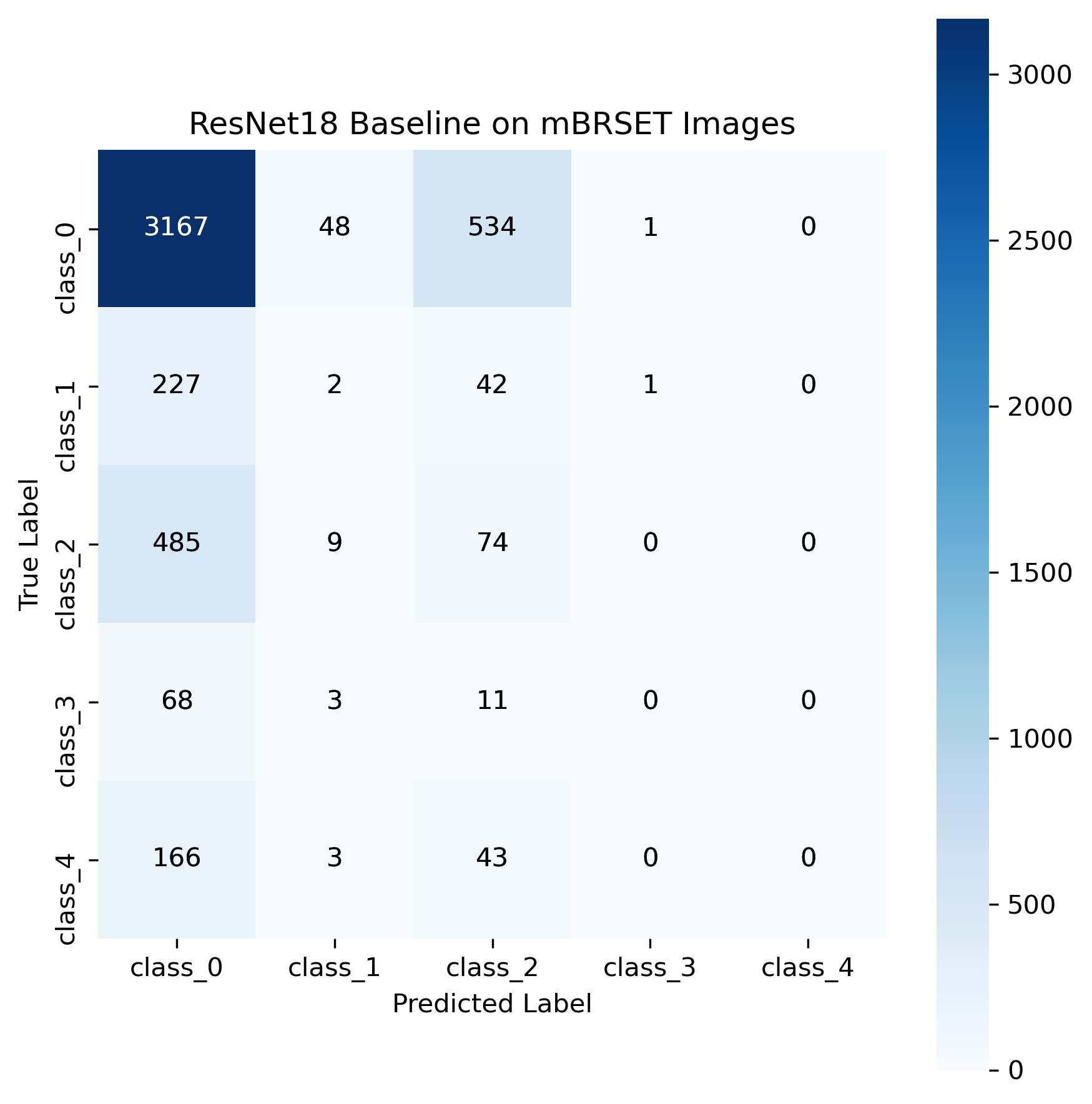}
    \caption{Confusion matrix illustrating the baseline ResNet18 model performance on mBRSET images.}
    \label{fig:baseline_cf}
\end{figure}

\begin{figure}[H]
    \centering    
    \includegraphics[width=0.8\linewidth]{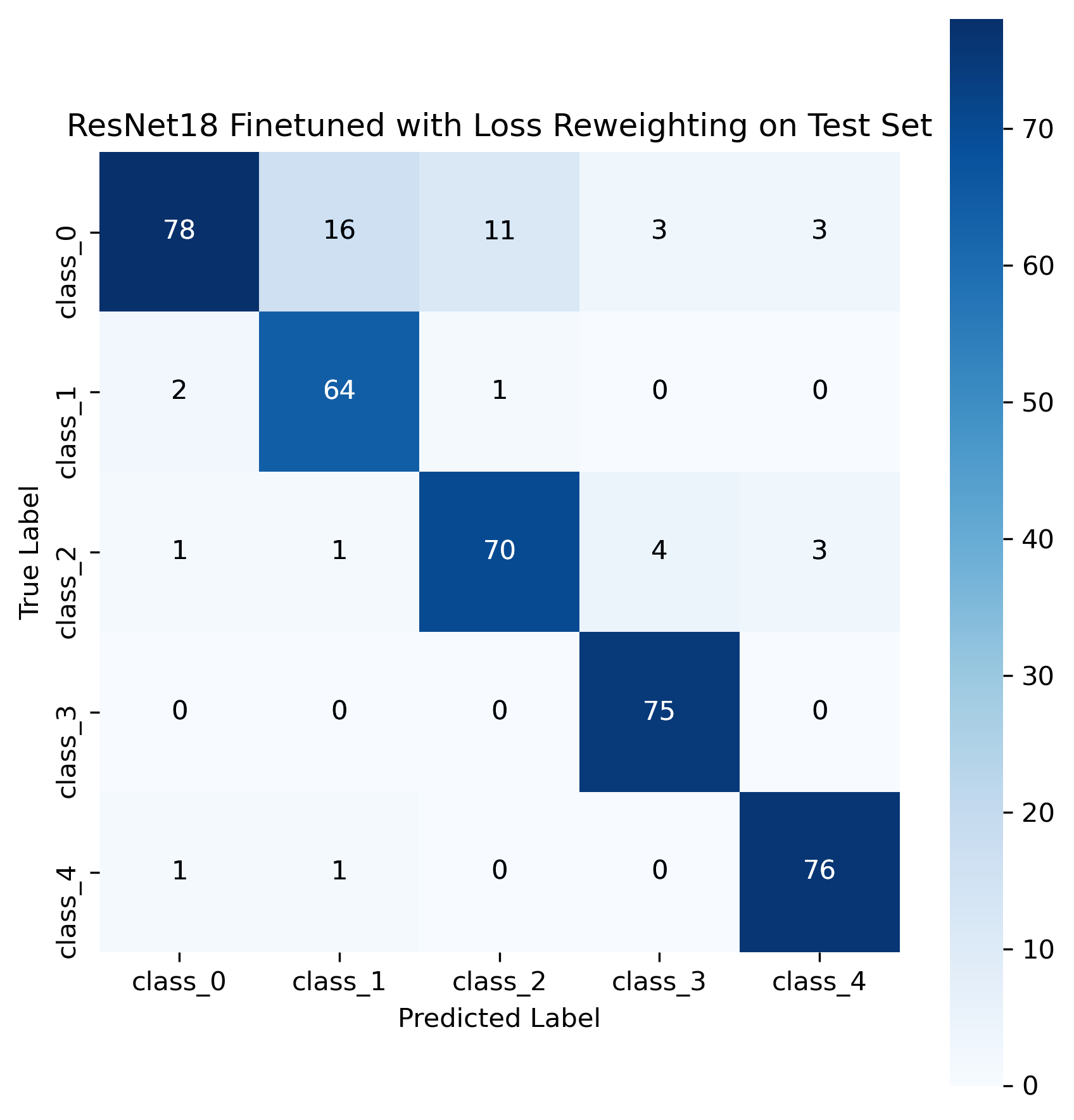}
    \caption{Confusion matrix illustrating the fine-tuned ResNet18 model performance on the held-out test set.}
    \label{fig:finetuned_cf}
\end{figure}

\paragraph{ResNet18}
The baseline ResNet18 model achieved an accuracy of 66.4\% and an averaged OvR AUROC of 0.534. The model was unable to accurately classify images labeled at more severe DR stages and demonstrated performance only marginally better than random chance (Figure \ref{fig:baseline_cf}). The performance drew our attention to the severe class imbalance present in this dataset. After fine-tuning, the model's performance increased substantially, reaching a final test accuracy of 88.5\% and averaged OvR AUROC of 0.986. The confusion matrix in Figure \ref{fig:finetuned_cf} shows that the model learned to distinguish between classes well, and particularly that our decision to use frequency weighting successfully mitigated the class imbalance.

\begin{figure}[H]
    \centering    
    \includegraphics[width=0.83\linewidth]{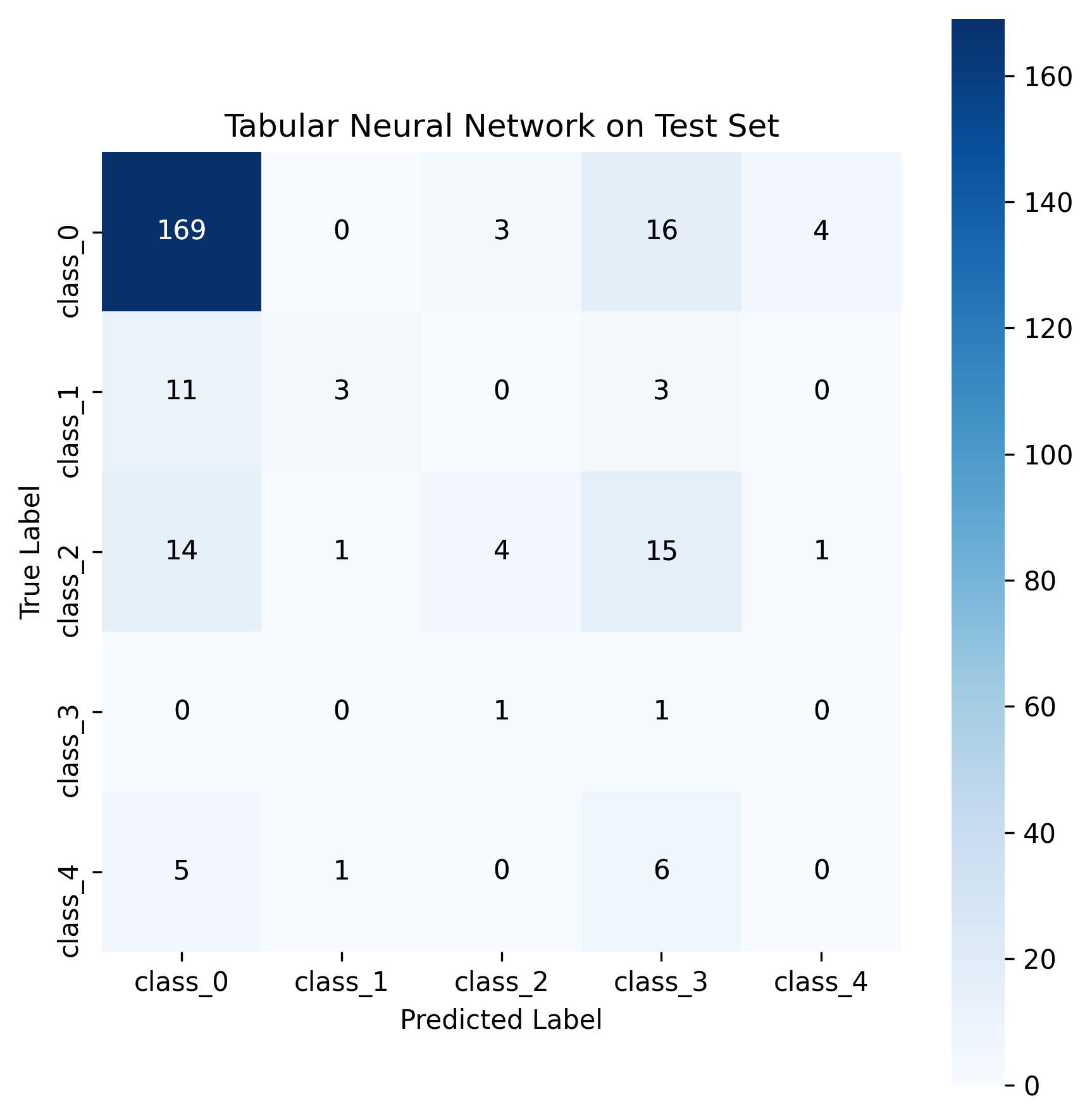}
    \caption{Confusion matrix illustrating the tabular neural network's predictions on the test dataset.}
    \label{fig:cm_tabular}
\end{figure}

\begin{figure}[H]
    \centering
    \includegraphics[width=0.9\linewidth]{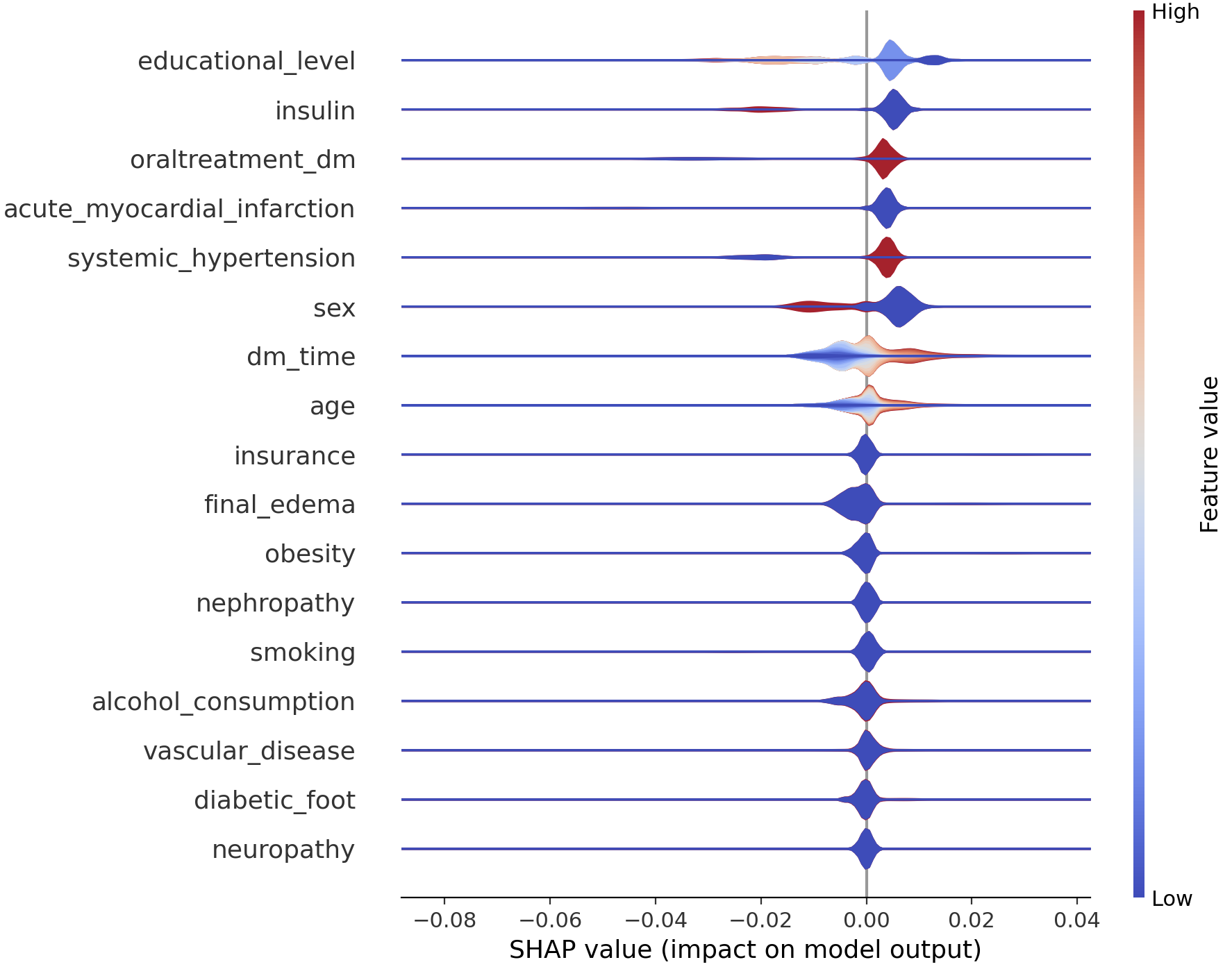}
    \caption{Representative violin plot illustrating SHAP scores for each of the 17 features. \texttt{Educational level}, \texttt{sex}, \texttt{dm\_time} (time since diabetes diagnosis), and \texttt{age} stand out for demonstrating nontrivial trends aligning with DR stage.}
    \label{fig:shap_tabular}
\end{figure}

\paragraph{Tabular Neural Network}
Our neural network trained solely on mBRSET's tabular features (socioeconomic factors and demographics) attained final test accuracy of 70.0\% and averaged OvR AUROC of 0.657. The confusion matrix suggests that the model may have learned some underlying relationships between tabular features and risk of DR (Figure \ref{fig:cm_tabular}), but we note that the relatively low AUROC here indicates a poor predictive relationship between these features and DR stage. Given that this is not time-series electronic health record data, but instead is a static descriptor, we expect that these features alone may not capture the nuanced DR stages well; we hypothesize that they may add a degree of additional context to retinal images in multimodal inference. SHAP values calculated across all five DR stages indicate \texttt{educational level}, \texttt{sex}, \texttt{time since diabetes diagnosis}, and \texttt{age} are four highly predictive tabular features (Figure \ref{fig:shap_tabular}).

\subsection{Multimodal fusion experimentation}
All three fusion methods showed similar performance on the classification task, as evidenced by their confusion matrices. We present the fully connected fusion results here (Figure \ref{fig:fc_layer}) and provide the remaining methods’ confusion matrices in the Supplementary Materials (Figures S1, S2). Concatenation performs slightly better than other methods looking at both training and testing accuracy and AUROC (Tables \ref{tab:train_fusion}, \ref{tab:test_fusion}). Particularly when compared to a unimodal baseline (a fine-tuned ResNet18 as described in Section \texttt{V}A), the multimodal model demonstrates substantially greater reliability in distinguishing between visually similar classes. Notably, both the concatenation and cross-attention fusion strategies resulted in zero instances of Classes 2, 3, and 4 being misclassified as Class 0 or 1; the same cannot be said for the unimodal baseline. This consistency in avoiding false negatives is crucial for the clinical implications of this classifier, where underestimating disease severity poses a greater risk than overestimating it. We continued to experiment with the three architectures on downstream tasks, namely generating three sets of embeddings as input to the deferral system.

\begin{figure}[H]
    \centering    
    \includegraphics[width=0.8\linewidth]{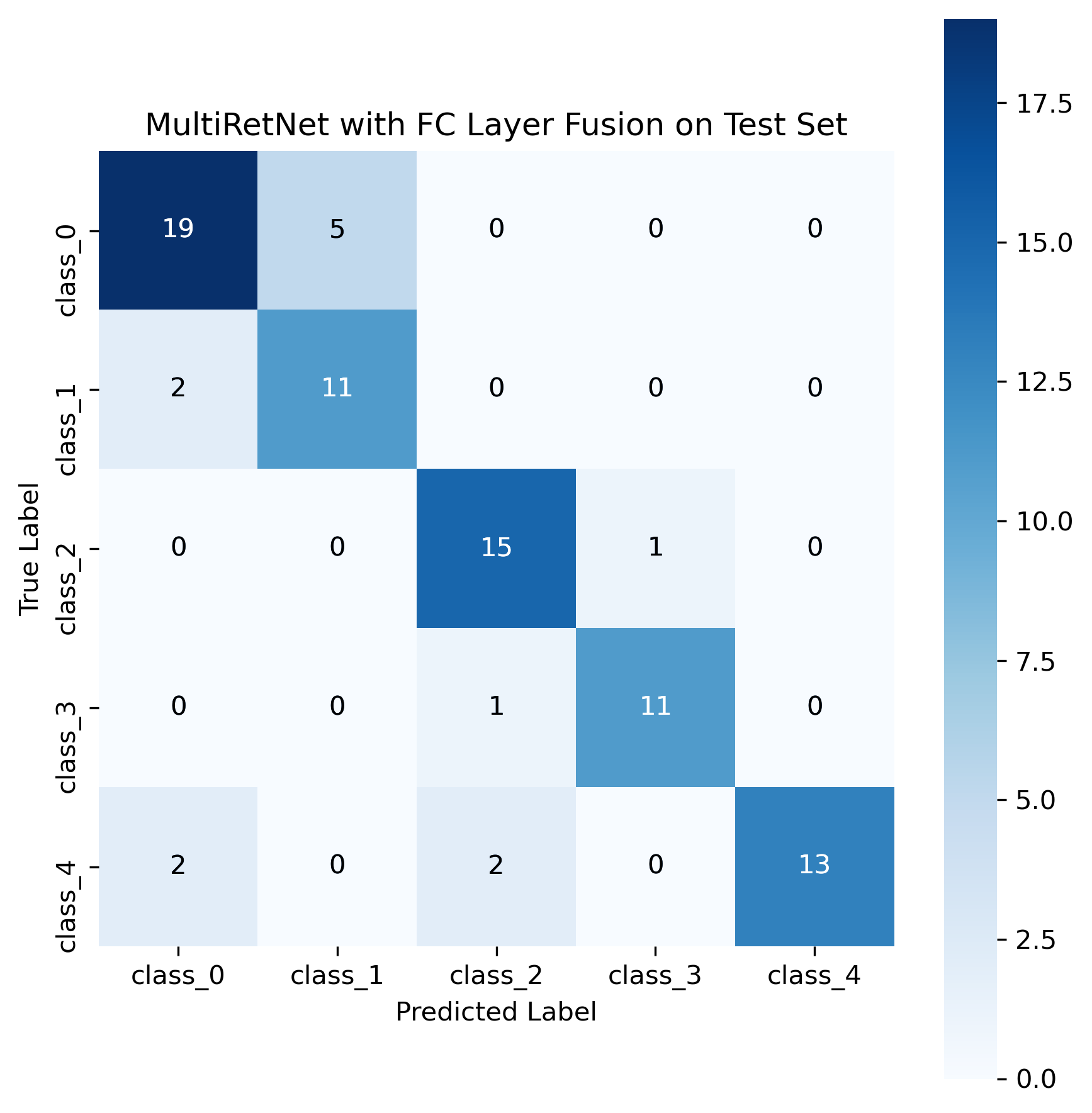}
    \caption{Confusion matrix illustrating the performance of the model using fully-connected layer-based fusion on the held-out test set.}
    \label{fig:fc_layer}
\end{figure}


\begin{table}[H]
    \centering
    \begin{tabular}{cccc}
         & Concatenate & FC Layer & Cross-Attention \\
        Accuracy & 0.929 & 0.881 & 0.905 \\
        AUROC & 0.993 & 0.987 & 0.988 \\
    \end{tabular}
    \caption{Average model performance across training folds on 5-fold cross validation.}
    \label{tab:train_fusion}
\end{table}

\begin{table}[H]
    \centering
    \begin{tabular}{cccc}
         & Concatenate & FC Layer & Cross-Attention \\
        Accuracy & 0.890 & 0.841 & 0.890 \\
        AUROC & 0.986 & 0.984 & 0.981 \\
    \end{tabular}
    \caption{Model performance on held-out test set.}
    \label{tab:test_fusion}
\end{table}
\end{multicols}

\begin{multicols}{2}

\subsection{Contrastive learning}
\begin{figure}[H]
    \centering
    \includegraphics[width=0.8\linewidth]{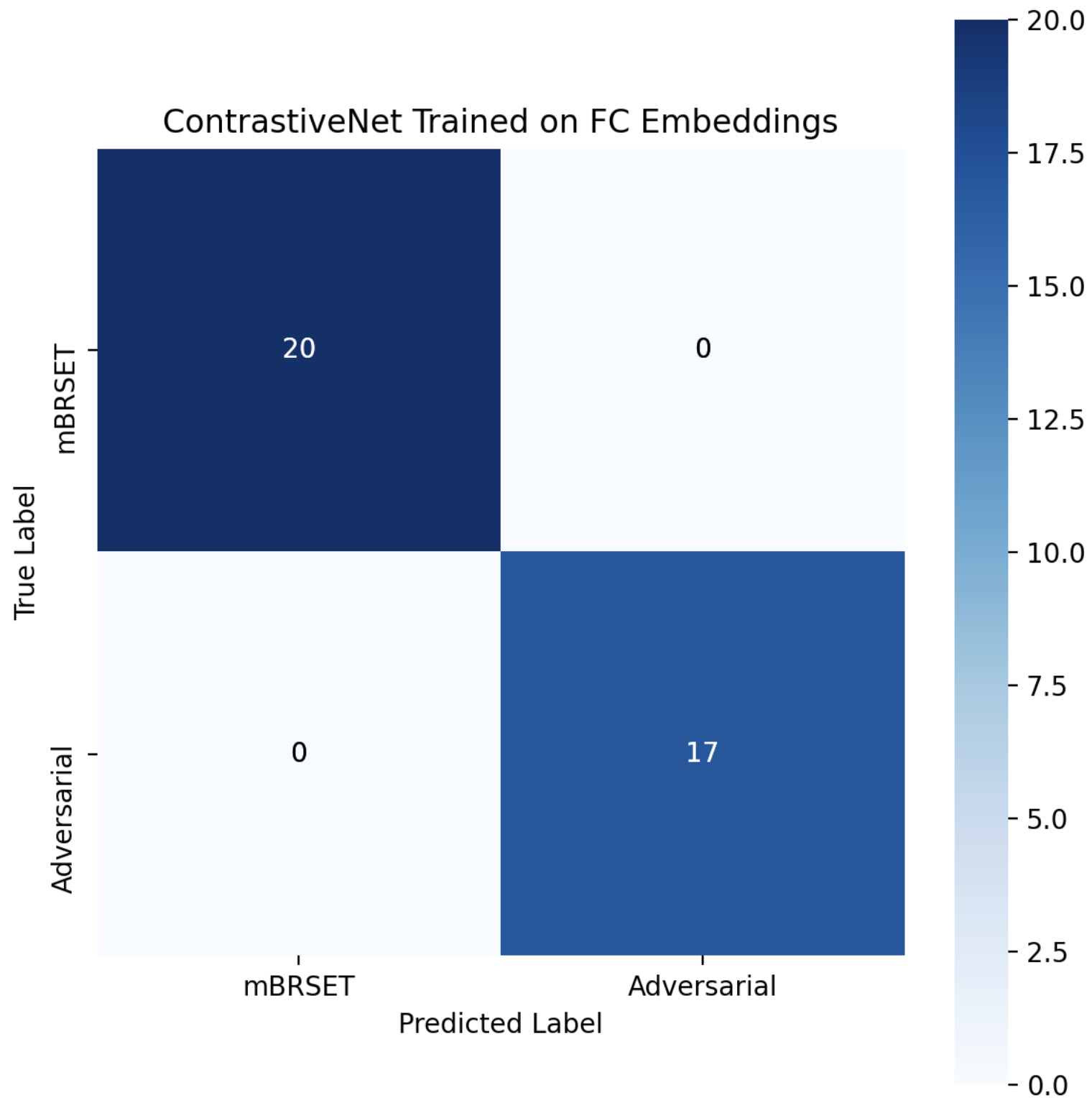} \\
    \includegraphics[width=\linewidth]{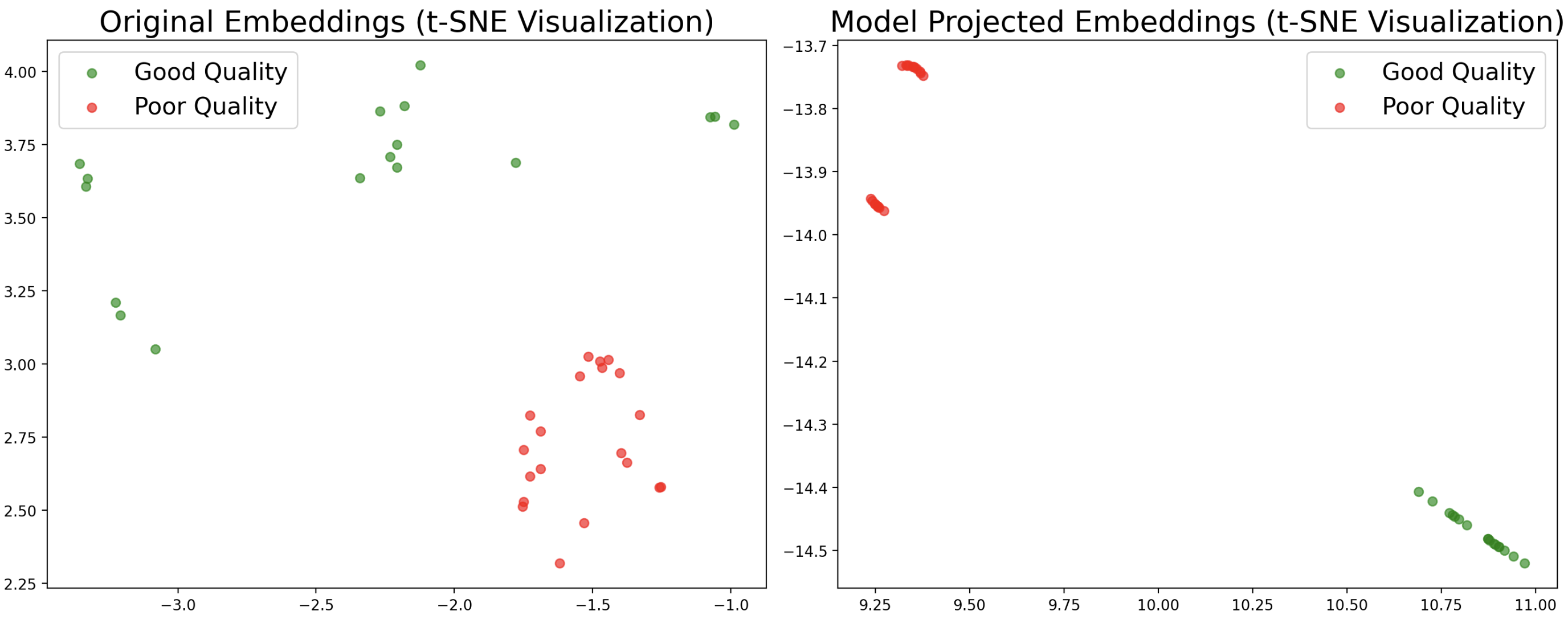}
    \caption{Confusion matrix and t-SNE illustrating test set performance of \texttt{ContrastiveNet} trained on embeddings generated through a fully-connected layer.}
    \label{fig:cm_contrastive_fc}
\end{figure}

We utilized our three methods of multimodal fusion to obtain three separate embeddings of mBRSET images as well as three separate embeddings of the adversarial images that we generated. We trained \texttt{ContrastiveNet} on the three appropriate pairs of embeddings, i.e., concatenated embeddings of mBRSET paired with concatenated embeddings of the adversarial images and similarly for the cross-attention embeddings and fully-connected embeddings. Quantification of model performance via confusion matrices as well as visual validation via t-SNE projections suggests, surprisingly, that the fully-connected embedding method is significantly superior for differentiating high-quality mBRSET images (along with associated tabular features) from low-quality adversarial images (Figure \ref{fig:cm_contrastive_fc}). In fact, both cross-attention embeddings and concatenated embeddings attain only 56.8\% and 59.5\% accuracy on the held-out test set, respectively (Figures S3, S4), whereas fully-connected embeddings attain perfect accuracy.

\section{Discussion}
Our results support our hypothesis that multimodal data is able to capture nuanced information that solely retinal information may not, particularly when comparing the performance of MultiRetNet against our unimodal baseline models. 
By comparing three methods of multimodal data fusion, we illustrate that utilizing newer, transformer-inspired architectures like cross attention captures at least as much information as concatenation. Even without particularly penalizing for false negatives in the classification task, cross-attention and concatenation avoided potentially harmful misclassifications of underestimating DR severity. It came as a surprise that concatenation slightly outperformed the deep learning fusion methods when evaluating MultiRetNet classification, but we hypothesize that this may be attributed to training on a dataset of only 328 images. Given a dataset orders of magnitude larger, or with additional methods to address class imbalance like upsampling, cross-attention may be advantageous over naive concatenation by enabling context-specific feature interactions and more expressive multimodal fusion. For the downstream task of training our deferral system, we note that the fully-connected method was able to capture differences in high-quality and low-quality images and achieved perfect separation of embeddings. We hypothesize that our embedding methods may follow a Goldilocks-like pattern, where simple concatenation is insufficient to enable meaningful multi-modal embedding; cross-attention may have overwritten meaningful information by performing too many complex operations; and a single fully-connected layer may balance simplicity with complexity to optimally generate a multimodal embedding.

\section{Conclusion}
By incorporating clinical features with retinal images, training on images taken without a high-quality fundus camera, and including a deferral system, MultiRetNet takes a human-centered approach to accurate, fast, and accessible diagnosis of diabetic retinopathy.

\section{Contributions and Acknowledgements}
J.S. worked on the image unimodal model, multimodal architecture design, multimodal fusion experimentation, and embedding generation.
K.S. worked on the tabular unimodal model, calculating SHAP scores, adversarial image generation, and deferral system architecture and experimentation.
The authors acknowledge the MIT Computer Vision class of Spring 2025 for motivating this research. The authors thank the MIT Office of Research Computing and Data (ORCD) for providing access to the Satori cluster for this work.

\section{Code and Data Availability}
The data used for this project is available from PhysioNet at \url{https://doi.org/10.13026/qxpd-1y65}. The code for this project is available at \url{https://github.com/jeannieshe/cv-dr}.
\end{multicols}

\newpage
\bibliographystyle{ieeetr}
\bibliography{cv_ref}

\begin{thebibliography}{10}

\bibitem{lu_divergent_2016}
Y.~Lu, ``Divergent {Perceptions} of {Barriers} to {Diabetic} {Retinopathy} {Screening} {Among} {Patients} and {Care} {Providers}, {Los} {Angeles}, {California}, 2014–2015,'' {\em Preventing Chronic Disease}, vol.~13, 2016.

\bibitem{cdc_vehss_2025}
CDC, ``{VEHSS} {Modeled} {Estimates}: {Prevalence} of {Diabetic} {Retinopathy} ({DR}),'' Feb. 2025.

\bibitem{uk_prospective_diabetes_study_group_tight_1998}
{UK Prospective Diabetes Study Group}, ``Tight blood pressure control and risk of macrovascular and microvascular complications in type 2 diabetes: {UKPDS} 38. {UK} {Prospective} {Diabetes} {Study} {Group},'' {\em BMJ (Clinical research ed.)}, vol.~317, pp.~703--713, Sept. 1998.

\bibitem{wong_increased_2016}
C.~W. Wong, E.~L. Lamoureux, C.-Y. Cheng, G.~C.~M. Cheung, E.~S. Tai, T.~Y. Wong, and C.~Sabanayagam, ``Increased {Burden} of {Vision} {Impairment} and {Eye} {Diseases} in {Persons} with {Chronic} {Kidney} {Disease} — {A} {Population}-{Based} {Study},'' {\em eBioMedicine}, vol.~5, pp.~193--197, Mar. 2016.
\newblock Publisher: Elsevier.

\bibitem{niederhauser_digital_2020}
O.~Niederhauser, ``Digital {Diagnostics}, formerly {IDx}, {Expands} {Global} {Impact} of {Healthcare} {Autonomous} {AI} with {Acquisition} of {3Derm} {Systems}, {Inc}.,'' Aug. 2020.

\bibitem{tufail_automated_2017}
A.~Tufail, C.~Rudisill, C.~Egan, V.~V. Kapetanakis, S.~Salas-Vega, C.~G. Owen, A.~Lee, V.~Louw, J.~Anderson, G.~Liew, L.~Bolter, S.~Srinivas, M.~Nittala, S.~Sadda, P.~Taylor, and A.~R. Rudnicka, ``Automated {Diabetic} {Retinopathy} {Image} {Assessment} {Software}: {Diagnostic} {Accuracy} and {Cost}-{Effectiveness} {Compared} with {Human} {Graders},'' {\em Ophthalmology}, vol.~124, pp.~343--351, Mar. 2017.

\bibitem{javitt_preventive_1994}
J.~C. Javitt, L.~P. Aiello, Y.~Chiang, F.~L. Ferris, J.~K. Canner, and S.~Greenfield, ``Preventive eye care in people with diabetes is cost-saving to the federal government. {Implications} for health-care reform,'' {\em Diabetes Care}, vol.~17, pp.~909--917, Aug. 1994.

\bibitem{dai_deep_2021}
L.~Dai, L.~Wu, H.~Li, C.~Cai, Q.~Wu, H.~Kong, R.~Liu, X.~Wang, X.~Hou, Y.~Liu, X.~Long, Y.~Wen, L.~Lu, Y.~Shen, Y.~Chen, D.~Shen, X.~Yang, H.~Zou, B.~Sheng, and W.~Jia, ``A deep learning system for detecting diabetic retinopathy across the disease spectrum,'' {\em Nature Communications}, vol.~12, p.~3242, May 2021.
\newblock Publisher: Nature Publishing Group.

\bibitem{li_review_2024}
Y.~Li, M.~El~Habib~Daho, P.-H. Conze, R.~Zeghlache, H.~Le~Boité, R.~Tadayoni, B.~Cochener, M.~Lamard, and G.~Quellec, ``A review of deep learning-based information fusion techniques for multimodal medical image classification,'' {\em Computers in Biology and Medicine}, vol.~177, p.~108635, July 2024.

\bibitem{beede_human-centered_2020}
E.~Beede, E.~Baylor, F.~Hersch, A.~Iurchenko, L.~Wilcox, P.~Ruamviboonsuk, and L.~M. Vardoulakis, ``A {Human}-{Centered} {Evaluation} of a {Deep} {Learning} {System} {Deployed} in {Clinics} for the {Detection} of {Diabetic} {Retinopathy},'' in {\em Proceedings of the 2020 {CHI} {Conference} on {Human} {Factors} in {Computing} {Systems}}, {CHI} '20, (New York, NY, USA), pp.~1--12, Association for Computing Machinery, Apr. 2020.

\bibitem{chen_simple_2020}
T.~Chen, S.~Kornblith, M.~Norouzi, and G.~Hinton, ``A {Simple} {Framework} for {Contrastive} {Learning} of {Visual} {Representations},'' July 2020.
\newblock arXiv:2002.05709 [cs].

\bibitem{wu_portable_2025}
C.~Wu, D.~Restrepo, L.~F. Nakayama, L.~Zago~Ribeiro, Z.~Shuai, N.~S. Barboza, M.~L.~V. Sousa, R.~D. Fitterman, A.~D.~A. Pereira, C.~V.~S. Regatieri, J.~A. Stuchi, F.~K. Malerbi, and R.~E. Andrade, ``A portable retina fundus photos dataset for clinical, demographic, and diabetic retinopathy prediction,'' {\em Scientific Data}, vol.~12, p.~323, Feb. 2025.
\newblock Publisher: Nature Publishing Group.

\bibitem{nakayama_mbrset_nodate}
L.~F. Nakayama, L.~Zago~Ribeiro, D.~Restrepo, N.~Santos~Barboza, R.~Dias~Fiterman, M.~l. Vieira~Sousa, A.~D.~A. Pereira, C.~Regatieri, F.~K. Malerbi, and R.~Andrade, ``{mBRSET}, a {Mobile} {Brazilian} {Retinal} {Dataset}.''

\bibitem{goldberger_physiobank_2000}
A.~L. Goldberger, L.~A. Amaral, L.~Glass, J.~M. Hausdorff, P.~C. Ivanov, R.~G. Mark, J.~E. Mietus, G.~B. Moody, C.~K. Peng, and H.~E. Stanley, ``{PhysioBank}, {PhysioToolkit}, and {PhysioNet}: components of a new research resource for complex physiologic signals,'' {\em Circulation}, vol.~101, pp.~E215--220, June 2000.

\end{thebibliography}

\end{document}


\maketitle
\vspace{-1.5cm}


\singlespacing
\setlength{\parskip}{6pt}
\setlength{\parindent}{0.5cm}

\setlength{\columnsep}{0.5cm}

    \begin{figure}[H]
    \centering    
    \includegraphics[width=0.8\linewidth]{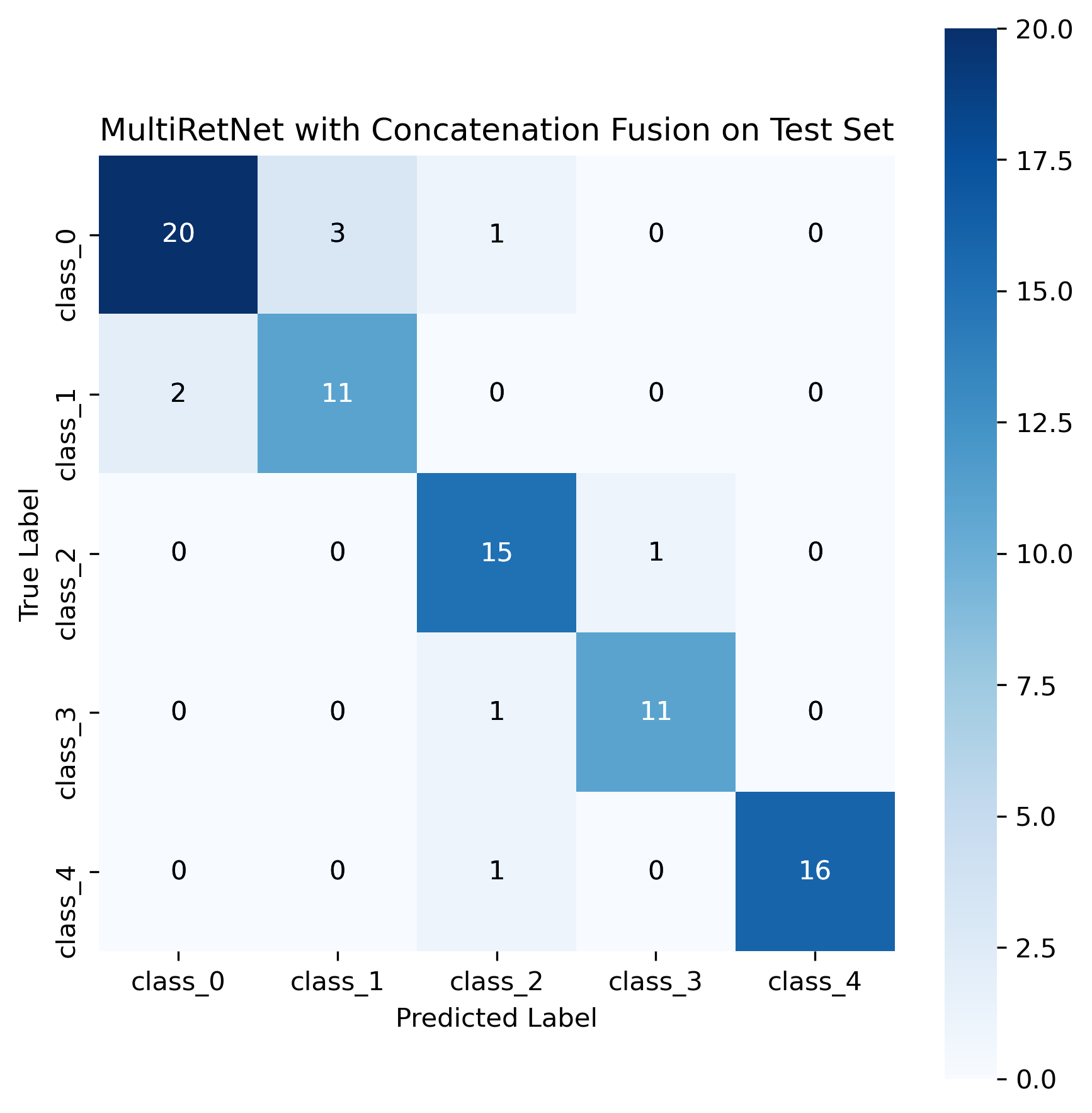}
    \caption{Confusion matrix illustrating the performance of the model using concatenation-based fusion on the held-out test set.}
    \label{fig:concatenation}
\end{figure}

\begin{figure}[H]
    \centering    
    \includegraphics[width=0.8\linewidth]{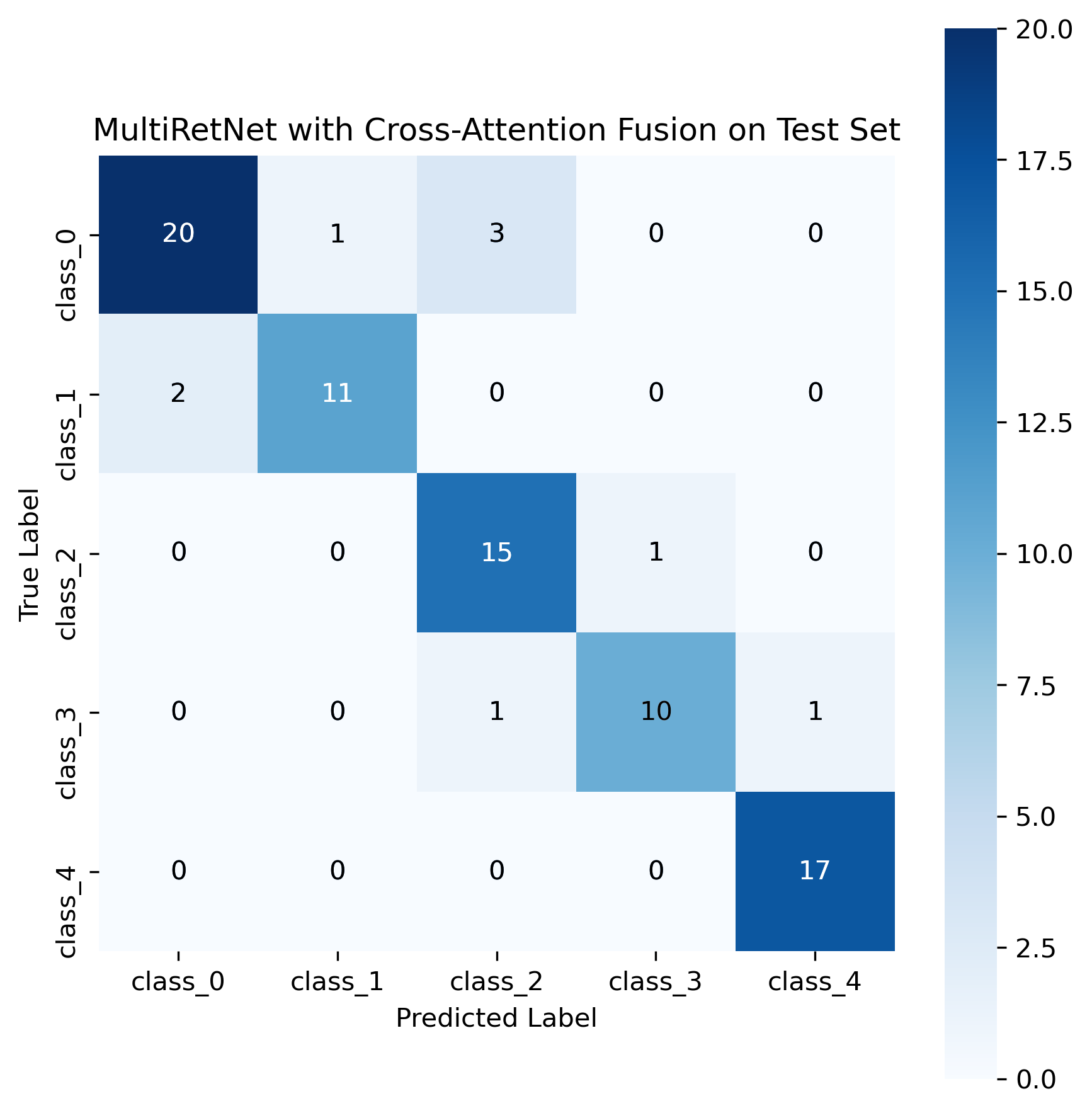}
    \caption{Confusion matrix illustrating the performance of the model using cross-attention-based fusion on the held-out test set.}
    \label{fig:cross_attention}
\end{figure}

\begin{figure}[H]
    \centering
    \includegraphics[width=0.8\linewidth]{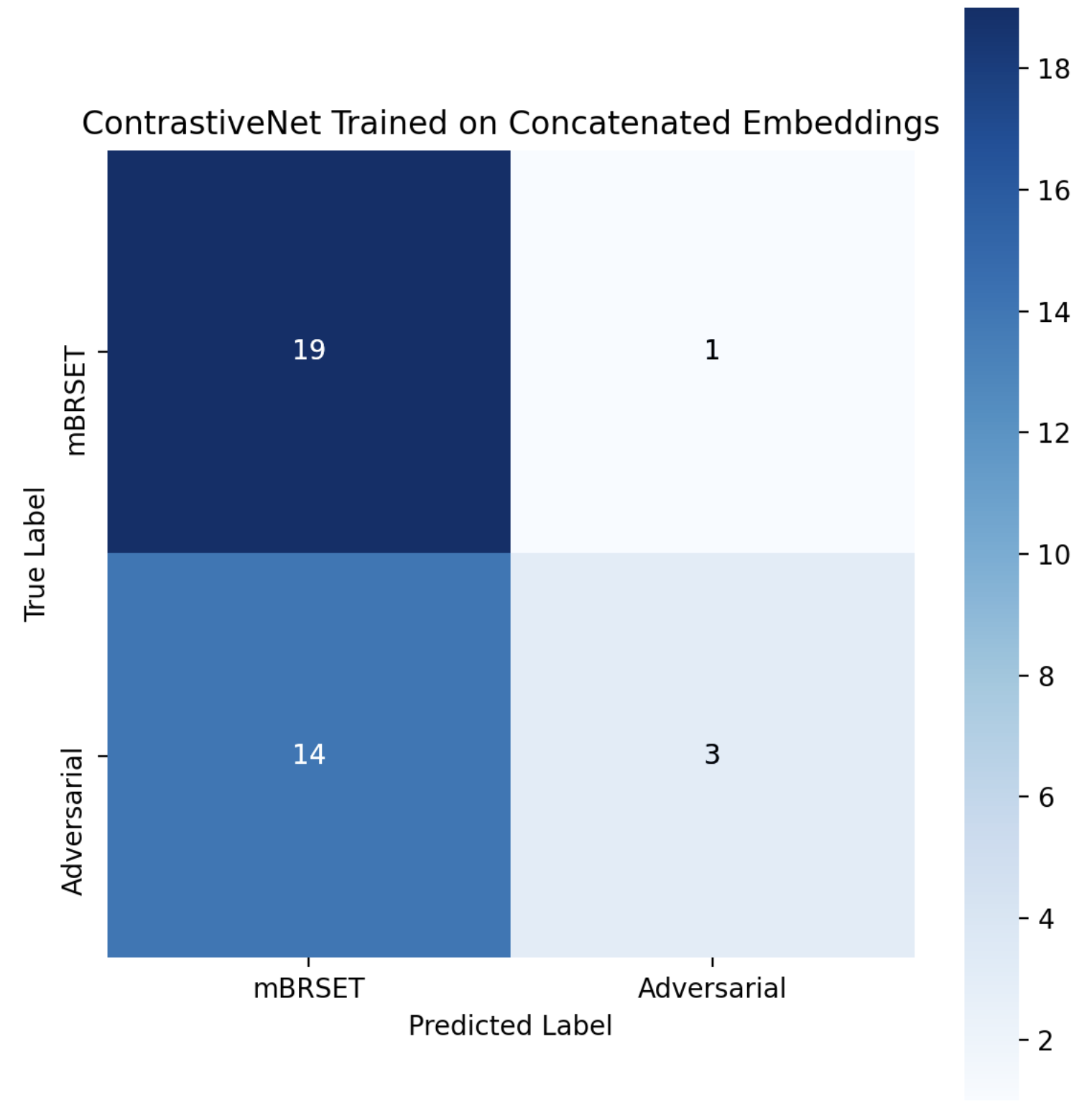} \\
    \includegraphics[width=0.8\linewidth]{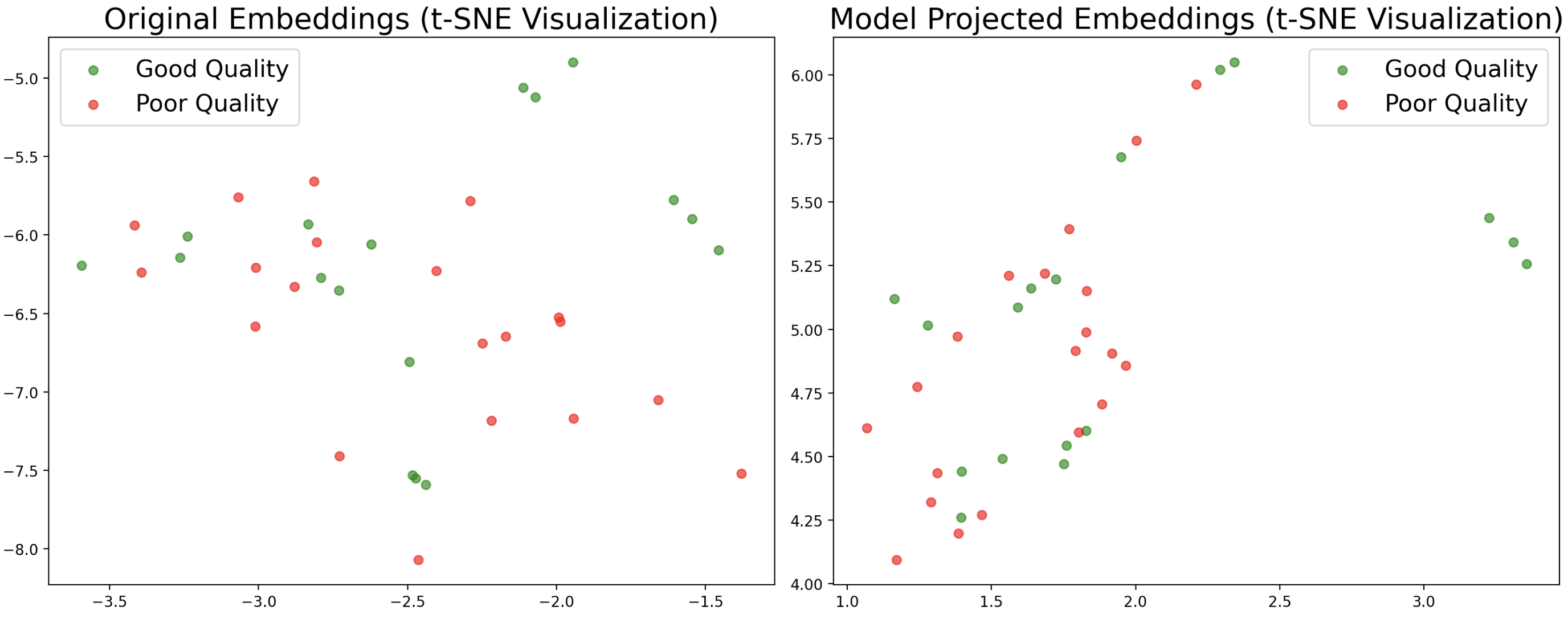}
    \caption{Confusion matrix and t-SNE illustrating test set performance of \texttt{ContrastiveNet} trained on embeddings generated through concatenation.}
    \label{fig:cm_contrastive_concat}
\end{figure}

\begin{figure}[H]
    \centering
    \includegraphics[width=0.8\linewidth]{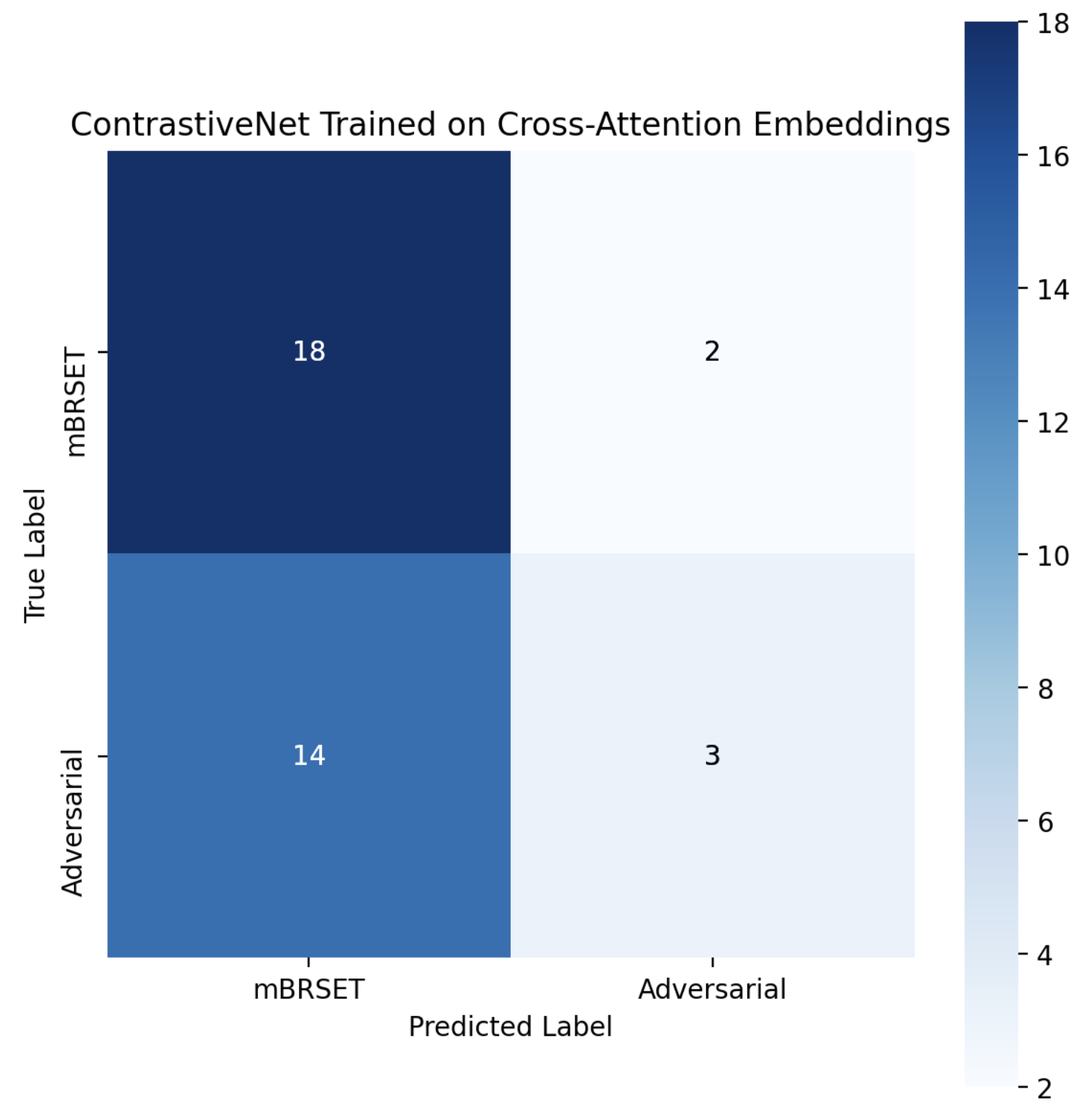} \\
    \includegraphics[width=0.8\linewidth]{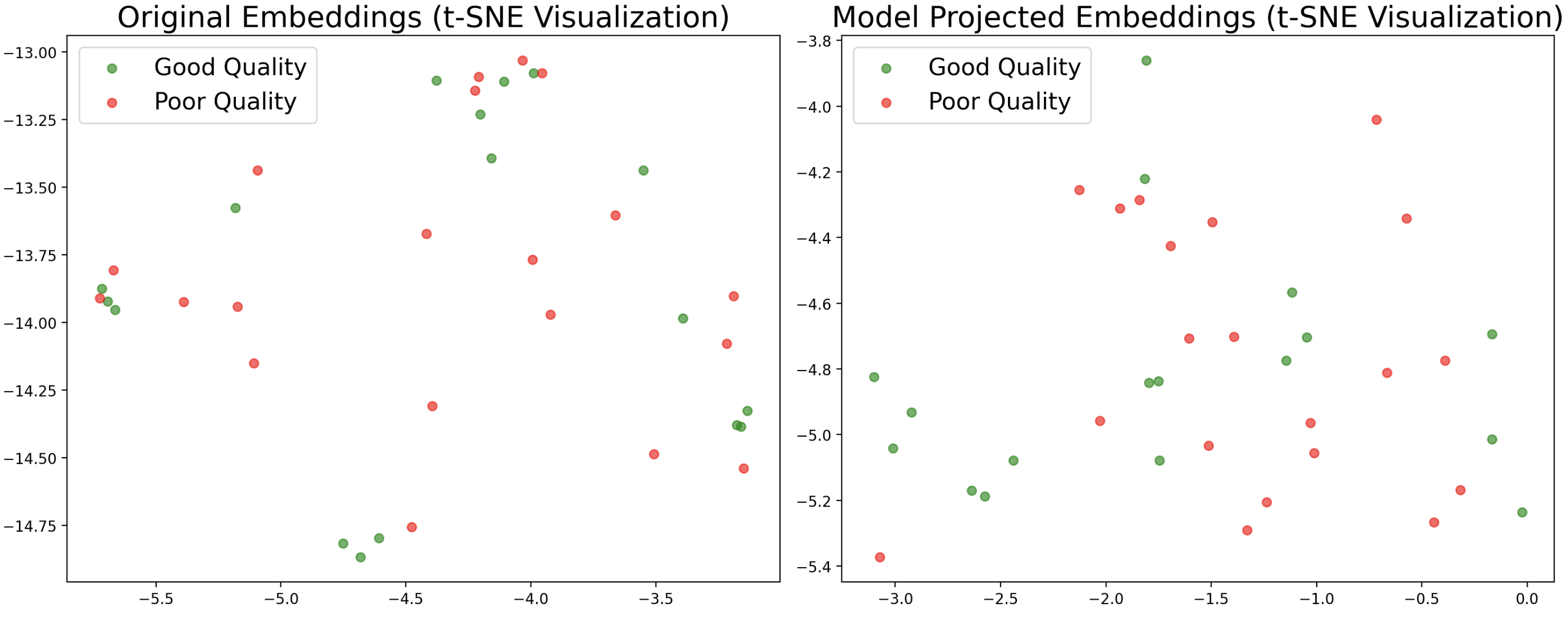}
    \caption{Confusion matrix and t-SNE illustrating test set performance of \texttt{ContrastiveNet} trained on embeddings generated through cross-attention.}
    \label{fig:cm_contrastive_crossattn}
\end{figure}